\documentclass[runningheads]{llncs}

\usepackage{cvconf}

\usepackage{cvconfabbrv}

\usepackage{graphicx}
\usepackage{booktabs}
\usepackage{algorithm}
\usepackage{algorithmic}

\usepackage[accsupp]{axessibility}

\usepackage[breaklinks,colorlinks,citecolor=confblue]{hyperref}

\usepackage{orcidlink}

\begin{document}

\title{Hypothesis Graph Refinement: Hypothesis-Driven Exploration with Cascade Error Correction for Embodied Navigation}

\titlerunning{Hypothesis Graph Refinement}

\author{
Peixin~Chen$^{1,2,\dagger}$ \and
Guoxi~Zhang$^{2,\dagger}$ \and
Jianwei~Ma$^{1,3,*}$ \and
Qing~Li$^{2,*}$
}

\authorrunning{P.~Chen et al.}

\institute{
$^1$Harbin Institute of Technology \qquad
$^2$Beijing Institute for General Artificial Intelligence \qquad
$^3$Peking University\\[4pt]
{\small $^\dagger$Equal contribution \qquad $^*$Corresponding author}\\[2pt]
}

\maketitle

\begin{abstract}
Embodied agents must explore partially observed environments while maintaining reliable long-horizon memory. Existing graph-based navigation systems improve scalability, but they often treat unexplored regions as semantically unknown, leading to inefficient frontier search. Although vision-language models (VLMs) can predict frontier semantics, erroneous predictions may be embedded into memory and propagate through downstream inferences, causing structural error accumulation that confidence attenuation alone cannot resolve. These observations call for a framework that can leverage semantic predictions for directed exploration while systematically retracting errors once new evidence contradicts them. We propose \textit{Hypothesis Graph Refinement} (HGR), a framework that represents frontier predictions as revisable hypothesis nodes in a dependency-aware graph memory. HGR introduces (1) \textit{semantic hypothesis module}, which estimates context-conditioned semantic distributions over frontiers and ranks exploration targets by goal relevance, travel cost, and uncertainty, and (2) \textit{verification-driven cascade correction}, which compares on-site observations against predicted semantics and, upon mismatch, retracts the refuted node together with all its downstream dependents. Unlike additive map-building, this allows the graph to contract by pruning erroneous subgraphs, keeping memory reliable throughout long episodes. We evaluate HGR on multimodal lifelong navigation (GOAT-Bench) and embodied question answering (A-EQA, EM-EQA). HGR achieves 72.41\% success rate and 56.22\% SPL on GOAT-Bench, and shows consistent improvements on both QA benchmarks. Diagnostic analysis reveals that cascade correction eliminates approximately 20\% of structurally redundant hypothesis nodes and reduces revisits to erroneous regions by $4.5\times$, with specular and transparent surfaces accounting for 67\% of corrected prediction errors. Code is available at \url{https://github.com/chenpppx/Hypothesis_Graph_Refinement}.

\end{abstract}

\section{Introduction}

Embodied agents operating in partially observed environments must decide where to explore next based on incomplete information. Classical frontier-based methods~\cite{etpnav2024, 3dmem2024} select boundaries between explored and unexplored space using geometric criteria, treating all frontiers as equally unknown. Recent work has begun leveraging vision-language models (VLMs) to predict what may lie beyond observed boundaries, enabling semantically informed frontier selection that prioritizes promising directions over exhaustive search~\cite{shah2023lm, majumdar2022zson, exploreeqa2024}. This shift from geometry-driven to semantics-driven exploration has demonstrated clear efficiency gains in short-horizon navigation tasks.

\begin{figure}[tbp]
  \centering
  \includegraphics[width=0.9\linewidth]{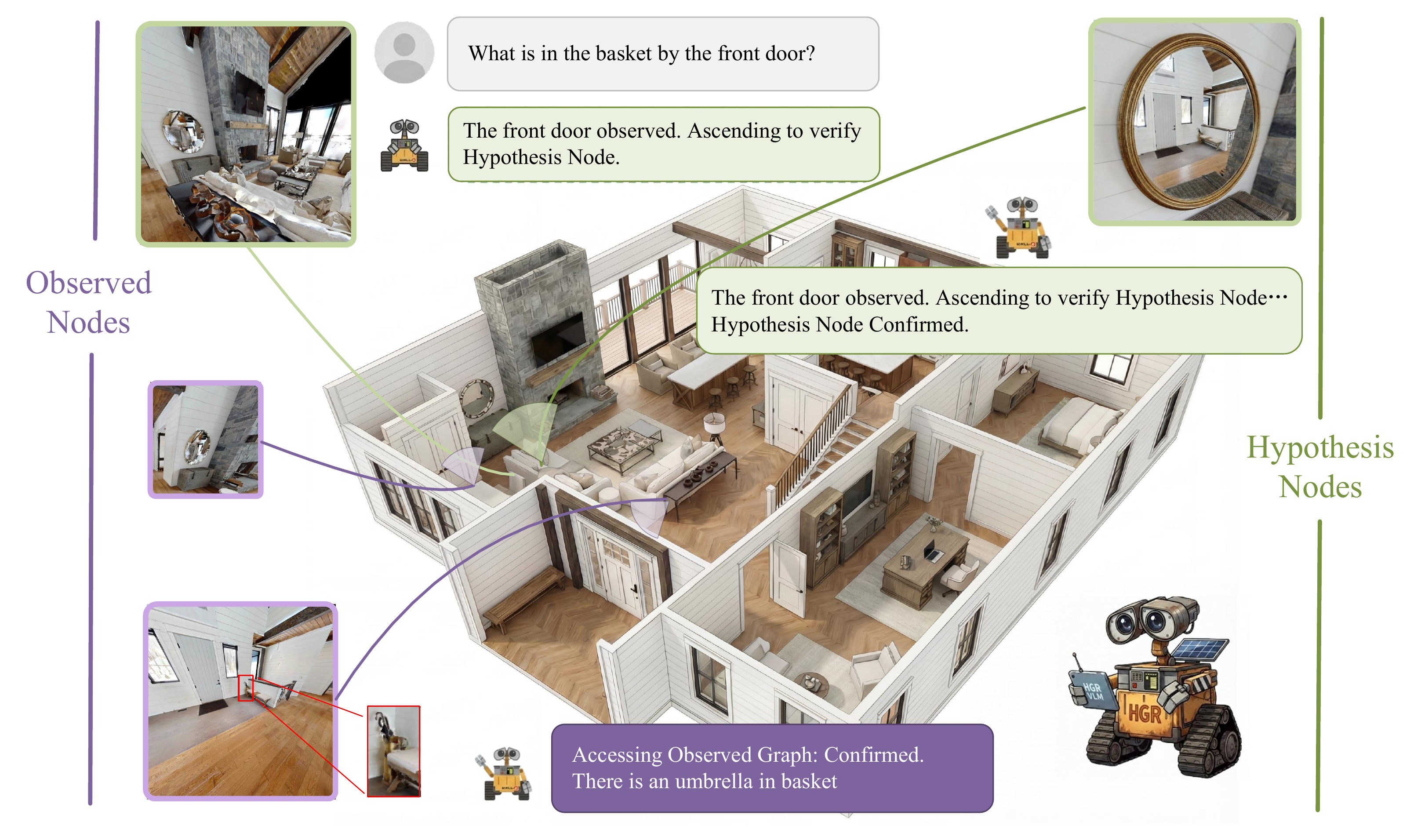}
  \caption{
  \textbf{Overview of Hypothesis Graph Refinement (HGR).}
  The hypothesis graph separates \textcolor[rgb]{0.4,0,0.6}{\textbf{observed nodes}} (purple, verified regions) from \textcolor[rgb]{0.3,0.5,0.3}{\textbf{hypothesis nodes}} (green, probabilistic frontier predictions), enabling a hypothesis-verification-correction cycle.
  \textbf{(Left)} Given the query ``\textit{What is inside the basket?}'', observed nodes provide confirmed scene context, while hypothesis nodes project semantic distributions onto unexplored frontiers, guiding the agent toward the most likely location of the target object.
  \textbf{(Right)} Upon arrival, the agent verifies each hypothesis against actual observations. If the prediction is confirmed, the hypothesis node transitions to an observed node; if refuted, cascade correction retracts the erroneous node and all its downstream dependents, preventing error propagation through the graph.
  }
  \label{fig:overview}
\end{figure}

However, VLM-based frontier predictions are not always reliable---visually ambiguous scenes such as mirror reflections or glass partitions can produce erroneous predictions that, once embedded in graph memory, propagate through chains of dependent hypotheses that inherit and amplify the original mistake. Existing approaches mitigate this through confidence attenuation, but the erroneous subgraph persists and continues to influence downstream decisions, causing cumulative degradation over long-horizon episodes.

We present \textbf{Hypothesis Graph Refinement} (HGR), a framework that jointly enables semantics-driven exploration and systematic error correction (Figure~\ref{fig:overview}). HGR maintains a \textit{hypothesis graph} that separates \textit{observed nodes} (verified regions) from \textit{hypothesis nodes} (probabilistic frontier predictions) and records their derivation relationships in a dependency DAG. A \textit{semantic hypothesis module} estimates semantic distributions over frontiers and ranks exploration targets by goal relevance, travel cost, and uncertainty. Upon visitation, a \textit{verification-driven cascade correction} mechanism compares actual observations against predictions; upon mismatch, it retracts the refuted node together with all downstream dependents, allowing the graph to contract by pruning erroneous subgraphs. Our contributions are:

\begin{itemize}
\item We introduce a hypothesis graph representation that distinguishes verified observations from revisable frontier predictions and tracks their dependencies, enabling structured hypothesis-driven exploration.

\item We propose a cascade correction mechanism that, upon detecting prediction failures, retracts the entire erroneous subgraph along the dependency DAG, preventing error accumulation over long episodes.

\item We demonstrate consistent improvements on GOAT-Bench~\cite{goatbench2024} (+3.31\% SR, +7.32\% SPL over 3D-Mem~\cite{3dmem2024}), A-EQA~\cite{openeqa2024}, and EM-EQA~\cite{openeqa2024}. Ablation and diagnostic analyses confirm that both mechanisms contribute meaningfully and that cascade correction substantially reduces error accumulation.
\end{itemize}

\section{Related Work}

\paragraph{Embodied Navigation and Exploration.} Embodied navigation evolves from reactive policies~\cite{anderson2018vision,ku2020room} to memory-augmented architectures~\cite{hong2021vln,deng2021evolving}. ETPNav~\cite{etpnav2024} introduces topological graphs for long-horizon navigation with Monte Carlo Tree Search for frontier evaluation. FILM~\cite{Min2021FILMFI} maintains episodic memories via transformer-based aggregation. Recent work explores VLM integration for open-vocabulary navigation~\cite{shah2023lm,majumdar2022zson}, but treats unexplored regions as lacking semantic information. Further advances include reinforcement fine-tuning for navigation~\cite{vlnr12025}, multi-turn active exploration~\cite{activevln2025}, and generative visual imagination~\cite{vista2025}. Concurrent work on commonsense-guided frontier exploration includes ESC~\cite{esc2023}, SCOPE~\cite{scope}, and ReVoLT~\cite{revolt2023}, which also leverage semantic priors to rank and select frontiers. However, these methods treat predictions as one-time scoring signals without recording derivation dependencies among them, so erroneous predictions persist in memory and may propagate to downstream inferences. HGR instead embeds each frontier prediction as a revisable hypothesis node in a dependency DAG, enabling systematic verification upon visitation and cascade correction that retracts an erroneous node together with all its dependents (see Appendix~\ref{app:related_comparison} for a detailed comparison).

\paragraph{3D Scene Representations.} Object-centric scene graphs~\cite{Armeni20193DSG,wald2020learning,conceptgraph2024} compress environments into nodes (objects) and edges (relationships). ConceptGraph~\cite{conceptgraph2024} pioneers open-vocabulary scene graphs via CLIP grounding. 3D-Mem~\cite{3dmem2024} employs memory snapshots capturing co-visible objects with spatial context. Recent extensions address contrastive pretraining~\cite{objectcentric3dsg2025}, retrieval-augmented reasoning~\cite{openworld3dsg2025}, metric-semantic graphs for embodied Q\&A~\cite{grapheqa2025}, and LLM-based semantic integration~\cite{3dgraphllm2025}. These representations handle spatial reasoning well but rely on soft probabilistic updates for error correction. HGR introduces dependency tracking and cascade correction to remove erroneous subgraphs rather than attenuating confidence scores.

\paragraph{Error Detection and Correction in Embodied Systems.} VLM prediction errors, where generated content is inconsistent with visual inputs, are studied extensively~\cite{blip2,sun-etal-2024-aligning}. Mitigation strategies include retrieval-augmented generation~\cite{Monaci2025RANaRN}, uncertainty quantification~\cite{kadavath2022language}, and self-consistency checks~\cite{manakul2023selfcheckgpt}. In recent robotics, COWS~\cite{gadre2023cows} uses contrastive learning and PIVOT~\cite{nasiriany2024pivot} employs verification modules.  ReLEP~\cite{relep2024}, HEAL~\cite{heal2025}, and InterleaveVLA~\cite{interleavevla2025} focus on output-level refinement, whereas HGR tracks dependency chains and removes erroneous subgraphs via cascade correction.

\section{Method}

\begin{figure*}[t]
  \centering
  \includegraphics[width=0.9\linewidth]{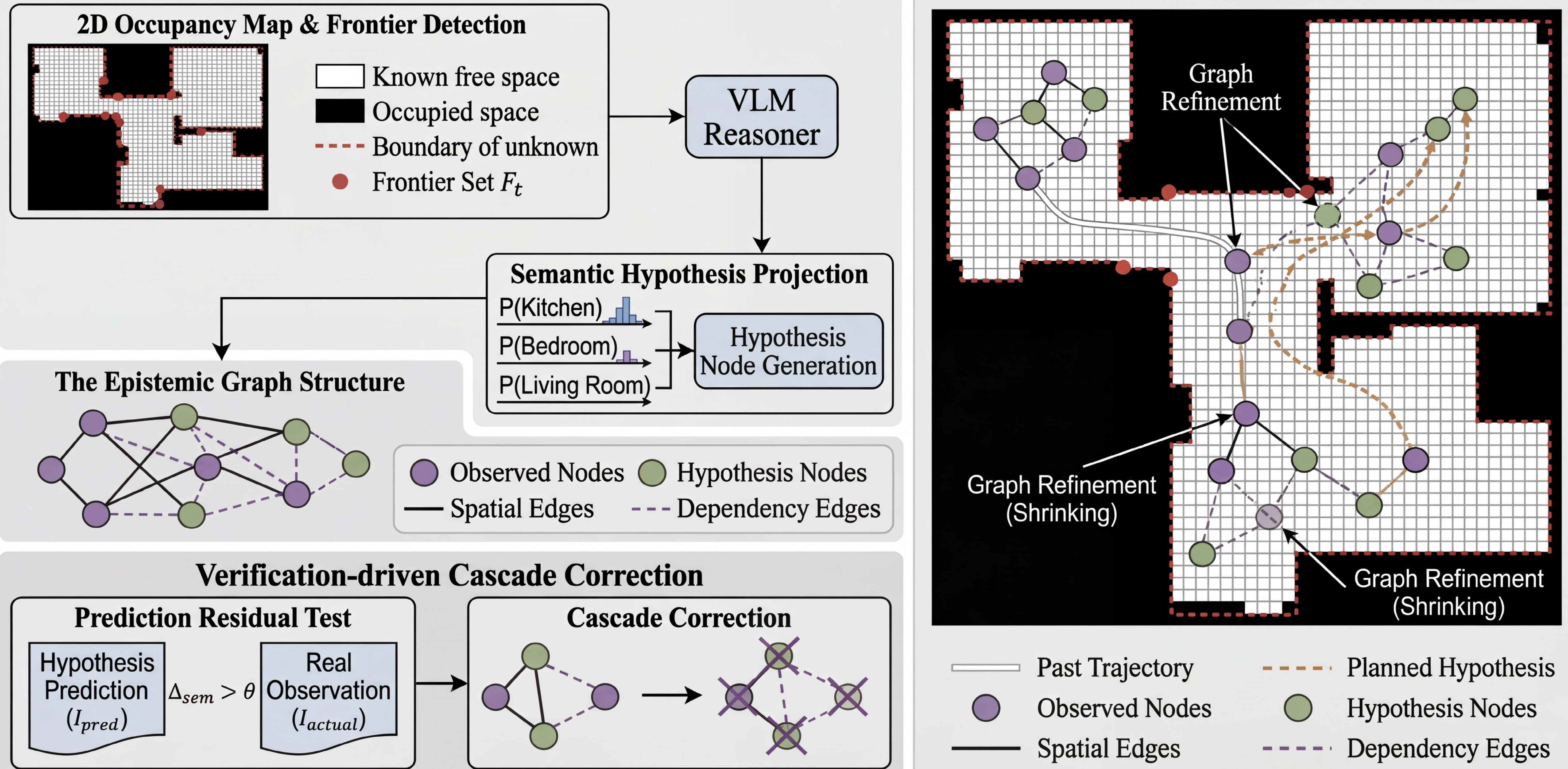}
  \caption{\textbf{Architecture of HGR.}
  \textbf{(Left)} Frontiers $\mathcal{F}_t$ detected from the occupancy map are fed to a VLM reasoner for \textit{semantic hypothesis module}, which estimates categorical distributions and generates \textcolor[rgb]{0.3,0.5,0.3}{\textbf{hypothesis nodes}} linked to \textcolor[rgb]{0.4,0,0.6}{\textbf{observed nodes}} via spatial and dependency edges. Upon visitation, \textit{cascade correction} compares predicted ($I_{\mathrm{pred}}$) and actual ($I_{\mathrm{actual}}$) semantics; if $\Delta_{\mathrm{sem}} > \theta$, the refuted node and all its dependents are removed.
  \textbf{(Right)} Running example on a floor plan. \textit{Graph Refinement} marks confirmed hypotheses promoted to observed nodes; \textit{Graph Refinement (Shrinking)} shows where cascade correction prunes erroneous subgraphs, contracting the graph.}
  \label{fig:architecture}
\end{figure*}

Frontier-based exploration faces two intertwined challenges: frontiers lack semantic cues to guide efficient exploration, yet VLM-based predictions risk embedding errors that propagate through dependent hypotheses over long horizons. HGR addresses both within a unified graph-based framework: the hypothesis graph separates verified observations from revisable frontier predictions and tracks their dependencies (Section~\ref{sec:graph_repr}), evolving through an incremental construction pipeline (Section~\ref{sec:incremental}). The semantic hypothesis module (Section~\ref{sec:projection}) tackles the first challenge by projecting goal-relevant distributions onto frontiers; verification-driven cascade correction (Section~\ref{sec:verification}) tackles the second by retracting erroneous subgraphs along the dependency DAG. Figure~\ref{fig:architecture} illustrates the overall architecture.

\subsection{Hypothesis Graph Representation}
\label{sec:graph_repr}

HGR maintains a hypothesis graph $\mathcal{G} = (\mathcal{V}, \mathcal{E}, \mathcal{D})$ that serves as the agent's persistent memory of the environment.

\paragraph{Node Types.} The node set $\mathcal{V} = \mathcal{V}^{\text{obs}} \cup \mathcal{V}^{\text{hyp}}$ separates two categories. \textit{Observed nodes} $v \in \mathcal{V}^{\text{obs}}$ represent physically visited regions, storing verified semantic labels, visual features, and detected object lists from direct perception. \textit{Hypothesis nodes} $v \in \mathcal{V}^{\text{hyp}}$ represent semantic predictions projected onto unexplored frontiers; each carries a probability distribution $P(\cdot \mid \mathcal{H}_t, f)$ over semantic categories and is marked as unverified.

\paragraph{Navigability Edges and Dependency DAG.} The edge set $\mathcal{E}$ encodes traversable paths for navigation. The directed acyclic graph $\mathcal{D}$ records derivation relationships among nodes: an edge $(v_p, v_c, \rho) \in \mathcal{D}$ indicates that hypothesis $v_c$ was derived from parent $v_p$ with confidence $\rho \in [0, 1]$. This structure enables systematic retraction: when a node is invalidated, all dependent descendants can be identified and removed along $\mathcal{D}$.

\subsection{Incremental Construction and Update}
\label{sec:incremental}

The hypothesis graph evolves incrementally as the agent explores through a three-phase cycle: \textit{hypothesis generation}, \textit{verification}, and \textit{update}. We outline each phase below; Sections~\ref{sec:projection} and~\ref{sec:verification} then detail the generation and verification mechanisms, respectively.

\paragraph{Phase 1: Frontier Discovery and Hypothesis Generation.} At each timestep $t$, given observation $(I_t, d_t, \mathbf{p}_t)$ (RGB, depth, pose), the system identifies geometric frontiers $\mathcal{F}_t$ from the occupancy map. For each new frontier $f_j$, a hypothesis node $v_j^{\text{hyp}}$ is generated via the semantic hypothesis module (Section~\ref{sec:projection}), added to $\mathcal{V}^{\text{hyp}}$, and linked in $\mathcal{D}$ to its parent observed node.

\paragraph{Phase 2: Hypothesis Verification.} When the agent reaches a hypothesis node $v_j^{\text{hyp}}$, it obtains actual observations and computes a prediction residual (Section~\ref{sec:verification}). Two outcomes follow: (1)~\textit{Confirmation}---prediction matches observation, and the node transitions to $\mathcal{V}^{\text{obs}}$ with updated labels; (2)~\textit{Refutation}---significant deviation triggers cascade correction, removing the node and all its dependents from $\mathcal{D}$.

\paragraph{Phase 3: Observed Node Update.} When revisiting an observed node, the system updates its object list and visual features from the latest perception. Appendix~\ref{app:dag_construction} provides full construction rules and worked examples.

\subsection{Semantic Hypothesis Module}
\label{sec:projection}

\begin{figure}[t]
  \centering
  \includegraphics[width=0.8\linewidth]{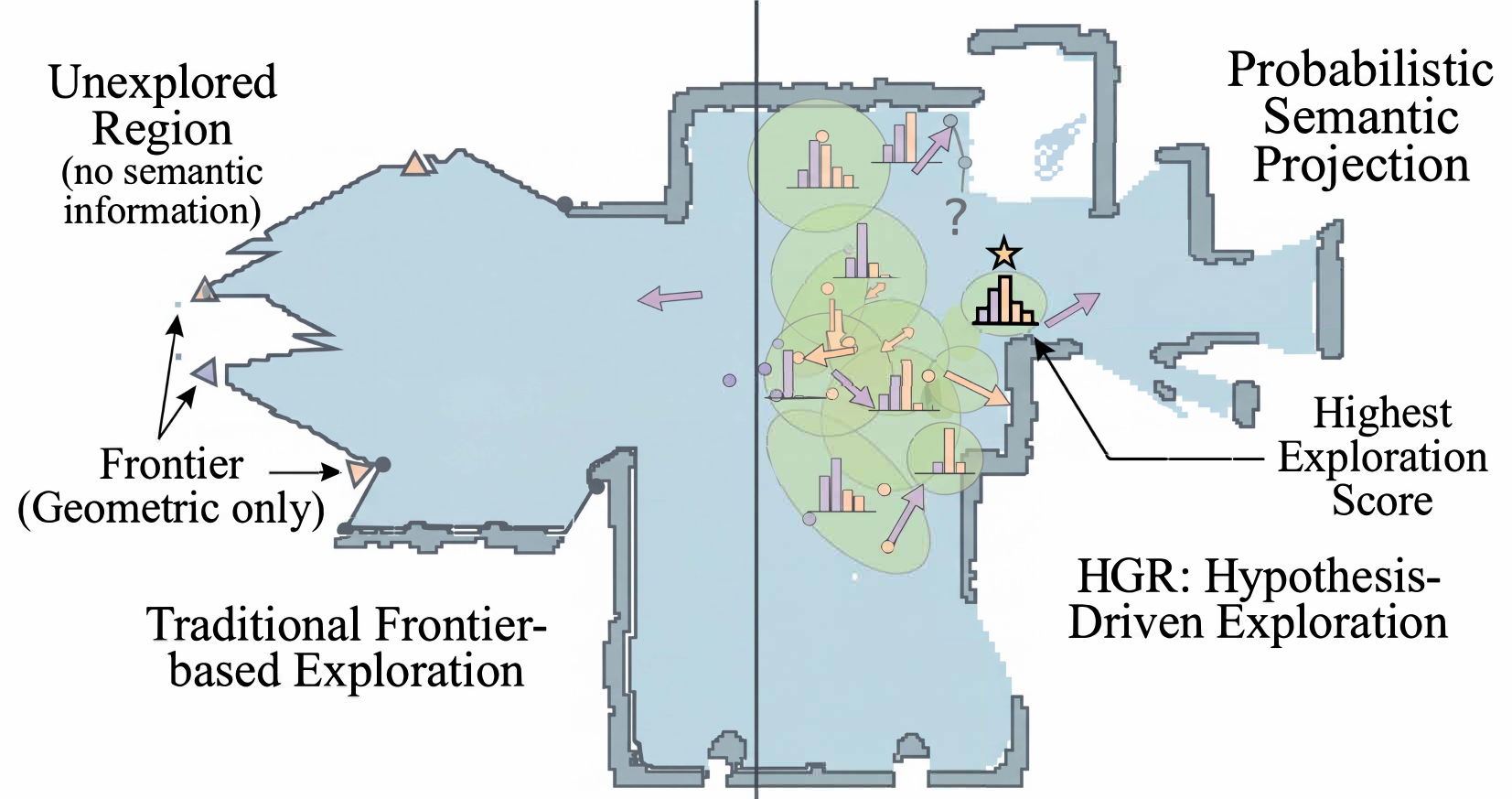}
  \caption{\textbf{Semantic Hypothesis Module.} Left: Traditional frontier representation treats unexplored regions as undifferentiated boundaries. Right: HGR projects probabilistic semantic distributions onto frontiers as hypothesis nodes, enabling goal-directed exploration.}
  \label{fig:semantic_projection}
\end{figure}

Rather than treating frontiers as undifferentiated boundaries (Figure~\ref{fig:semantic_projection}, left), HGR projects probabilistic semantic distributions onto each frontier, producing hypothesis nodes with estimated categorical distributions that enable goal-directed exploration (Figure~\ref{fig:semantic_projection}, right).

\paragraph{Hypothesis Node Generation.} Given observation history $\mathcal{H}_t = \{(I_i, d_i, \mathbf{p}_i)\}_{i=1}^t$ and the frontier set $\mathcal{F}_t = \{f_j\}$ identified in Phase~1, each frontier $f_j$ is assigned a semantic distribution over categories $\mathcal{C} = \{c_1, \ldots, c_K\}$:
\begin{equation}
P(c_k \mid \mathcal{H}_t, f_j) \propto \phi_{\text{VLM}}(c_k \mid I_j^{\text{frontier}}, \mathcal{O}_{\text{local}}),
\end{equation}
where $I_j^{\text{frontier}}$ is the visual observation toward frontier $f_j$, $\mathcal{O}_{\text{local}}$ denotes detected objects within spatial radius $r_{\text{context}}$, and $\phi_{\text{VLM}}$ uses the VLM's world knowledge to estimate semantic categories from partial visual cues and spatial context.

\paragraph{Exploration Scoring.} To convert frontier exploration into goal-directed hypothesis testing, we define an exploration score for each frontier $f_j$ relative to navigation goal $g$:
\begin{equation}
S(f_j; g) = \underbrace{P(c_g \mid \mathcal{H}_t, f_j)}_{\text{goal alignment}} - \underbrace{\lambda_d \cdot d(f_j, \mathbf{p}_t)}_{\text{travel cost}} + \underbrace{\lambda_h \cdot H[P(\cdot \mid \mathcal{H}_t, f_j)]}_{\text{uncertainty bonus}},
\label{eq:exploration_score}
\end{equation}
where $c_g$ is the target semantic category, $d(\cdot, \cdot)$ measures geodesic distance, and $H[\cdot]$ is Shannon entropy. The three terms respectively prioritize goal-matching frontiers, penalize distant ones, and encourage visiting ambiguous regions with high information gain.

\subsection{Verification and Cascade Correction}
\label{sec:verification}

\begin{figure*}[t]
  \centering
  \includegraphics[width=0.9\linewidth]{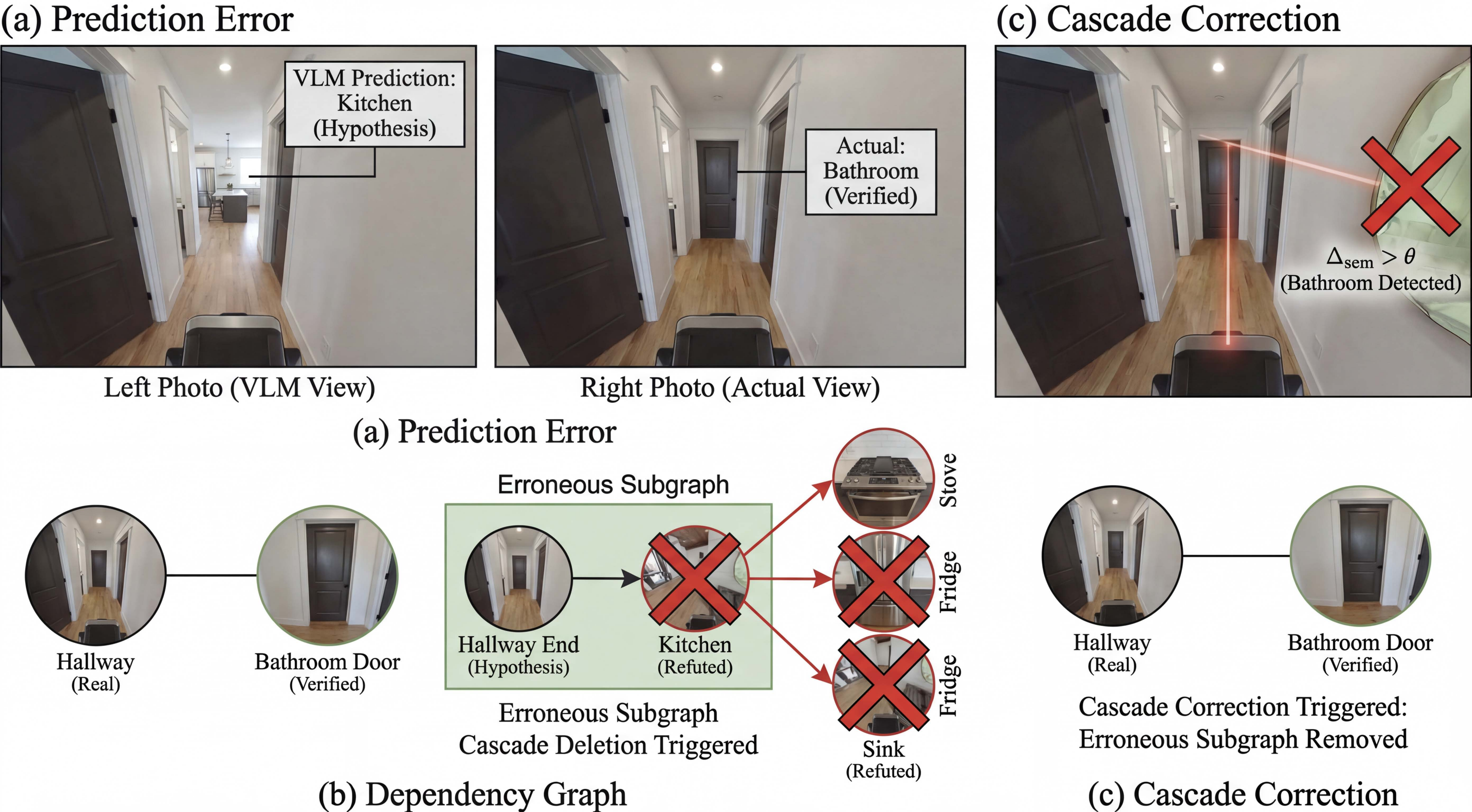}
  \caption{\textbf{Cascade Correction Example.} A VLM misidentifies a mirror reflection as a bedroom entrance, generating hypothesis nodes for inferred furniture. Upon reaching the mirror and detecting a prediction violation (residual $> \theta_{\text{refute}}$), the system traces the dependency DAG and removes the entire erroneous subgraph, including all descendant hypothesis nodes.}
  \label{fig:cascade_deletion}
\end{figure*}

When a hypothesis node is refuted in Phase~2, HGR must not only remove the erroneous node but also retract all downstream nodes derived from it. This section details the prediction residual test that triggers refutation and the cascade correction that propagates it through $\mathcal{D}$. Figure~\ref{fig:cascade_deletion} illustrates a representative case: a VLM misidentifies a mirror reflection as a room entrance, generating descendant hypothesis nodes for inferred furniture; upon detecting the semantic violation, cascade correction traces $\mathcal{D}$ and removes the entire erroneous subgraph.

\paragraph{Prediction Residual Test.} Upon reaching a hypothesis node $v_j^{\text{hyp}}$, the agent obtains actual perception $(I_j^{\text{actual}}, \mathcal{O}_j^{\text{actual}})$. The prediction residual quantifies the discrepancy between predicted and observed semantics:
\begin{equation}
\Delta_{\text{sem}}(v_j^{\text{hyp}}) = \omega_c \cdot \Delta_c + \omega_f \cdot \Delta_f + \omega_o \cdot \Delta_o,
\label{eq:semantic_residual}
\end{equation}
where:
\begin{align}
\Delta_c &= 1 - \max_{k} P(c_k \mid \mathcal{H}_t, f_j) \cdot \mathbb{1}[c_k = c_j^{\text{actual}}], \label{eq:residual_category} \\
\Delta_f &= 1 - \text{sim}_{\text{CLIP}}(\text{Enc}(I_j^{\text{frontier}}), \text{Enc}(I_j^{\text{actual}})), \label{eq:residual_feature} \\
\Delta_o &= 1 - \frac{|\mathcal{O}_j^{\text{pred}} \cap \mathcal{O}_j^{\text{actual}}|}{|\mathcal{O}_j^{\text{pred}} \cup \mathcal{O}_j^{\text{actual}}|}, \label{eq:residual_object}
\end{align}
combining category mismatch $\Delta_c$, visual feature divergence $\Delta_f$ via CLIP embeddings, and object-level Jaccard dissimilarity $\Delta_o$. Weights are set to $\omega_c = 0.4, \omega_f = 0.3, \omega_o = 0.3$ based on held-out validation; $\theta_{\text{refute}} = 0.5$ (see Appendix~\ref{app:residual_details} for per-term ablation and threshold sensitivity). A hypothesis node is \textit{refuted} when $\Delta_{\text{sem}} > \theta_{\text{refute}}$.

\paragraph{Cascade Correction.} Upon refuting a hypothesis node $v_p$, the system traverses the dependency DAG $\mathcal{D}$ to identify and remove all transitive descendants:

\begin{algorithm}[t]
\caption{Cascade Correction}
\label{alg:cascade_correction}
\begin{algorithmic}[1]
\STATE \textbf{Input:} Refuted node $v_p$, dependency DAG $\mathcal{D}$
\STATE \textbf{Output:} Set of removed nodes $\mathcal{V}_{\text{del}}$
\STATE Initialize queue $Q \leftarrow \{v_p\}$, $\mathcal{V}_{\text{del}} \leftarrow \emptyset$
\WHILE{$Q \neq \emptyset$}
    \STATE $v_{\text{curr}} \leftarrow Q.\text{dequeue}()$
    \STATE $\mathcal{V}_{\text{del}} \leftarrow \mathcal{V}_{\text{del}} \cup \{v_{\text{curr}}\}$
    \FOR{each child $v_c$ where $(v_{\text{curr}}, v_c, \rho) \in \mathcal{D}$}
        \STATE $Q.\text{enqueue}(v_c)$
    \ENDFOR
\ENDWHILE
\STATE Remove $\mathcal{V}_{\text{del}}$ from $\mathcal{G}$; remove corresponding edges from $\mathcal{E}$ and $\mathcal{D}$
\RETURN $\mathcal{V}_{\text{del}}$
\end{algorithmic}
\end{algorithm}

Invalidating a single hypothesis thus removes all downstream nodes inferred from the false premise. Unlike confidence attenuation, cascade correction eliminates erroneous subgraphs entirely, so the hypothesis graph may \textit{shrink} over time---a non-monotonic property distinguishing HGR from conventional additive map construction.

\section{Experiments}

We evaluate HGR on three complementary benchmarks to validate two hypotheses: (H1)~semantic hypothesis module improves exploration efficiency, and (H2)~cascade correction prevents cumulative errors from degrading performance.

\subsection{Experimental Setup}

\paragraph{Benchmarks.}
\textbf{(1) GOAT-Bench}~\cite{goatbench2024}: Multimodal lifelong navigation across 360 scenes requiring sequential navigation to multiple targets specified via category labels, language descriptions, or reference images. Episodes span 100+ steps, emphasizing sustained autonomy. This is our primary benchmark.
\textbf{(2) A-EQA}~\cite{openeqa2024}: Active Embodied Q\&A comprising 557 questions across 63 HM3D scenes. Agents must explore unknown environments to answer open-ended questions spanning object recognition, spatial understanding, and functional reasoning.
\textbf{(3) EM-EQA}~\cite{openeqa2024}: Episodic Memory Q\&A with 1,600+ questions from 152 ScanNet and HM3D scenes. Given pre-explored trajectories, agents construct scene memory and answer questions without further exploration, isolating memory quality from exploration strategy.

\paragraph{Metrics.} \textbf{Success Rate (SR)} measures goal achievement within distance thresholds ($1.0$m for navigation, answer correctness for Q\&A). \textbf{Success weighted by Path Length (SPL)} penalizes inefficient exploration: $\text{SPL} = \frac{1}{N} \sum_{i=1}^N S_i \cdot \frac{l_i^*}{\max(l_i^*, l_i)}$. For A-EQA, \textbf{LLM-Match} evaluates answer quality via GPT-4 scoring (0--100 scale).

\paragraph{Baselines.} Comparisons include \textbf{3D-Mem}~\cite{3dmem2024},  and \textbf{ConceptGraph}~\cite{conceptgraph2024} with frontier-based exploration, \textbf{Explore-EQA}~\cite{exploreeqa2024}, and ablated variants. For fair comparison, all baselines are re-implemented within our framework using the same GPT-4o VLM, Habitat simulator, low-level controller, and step budget. ConceptGraph and 3D-Mem retain their original graph construction and update logic but use GPT-4o in place of their original semantic modules. See Appendix~\ref{app:baseline_details} for full re-implementation details.

\subsection{Main Results: GOAT-Bench}

\paragraph{Lifelong Navigation Performance.} Table~\ref{tab:goat_main} presents results on GOAT-Bench. HGR achieves 72.41\% SR and 56.22\% SPL on the validation subset (278 subtasks), outperforming the strongest baseline 3D-Mem (69.1\%/48.9\%) by +3.31\% SR and +7.32\% SPL.

\begin{table}[t]
\caption{\textbf{Performance on GOAT-Bench.} HGR achieves the highest success rate and path efficiency on both the full validation set (2,780 subtasks) and the evaluation subset (278 subtasks).}
\label{tab:goat_main}
\centering
\begin{tabular}{lcccc}
\toprule
& \multicolumn{2}{c}{Full Validation Set} & \multicolumn{2}{c}{Subset} \\
\cmidrule(lr){2-3} \cmidrule(lr){4-5}
Method & SR $\uparrow$ & SPL $\uparrow$ & SR $\uparrow$ & SPL $\uparrow$ \\
\midrule
ConceptGraph~\cite{conceptgraph2024} & 61.2 & 44.3 & 67.8 & 48.1 \\
3D-Mem w/o memory~\cite{3dmem2024} & 58.6 & 38.5 & 66.2 & 44.1 \\
3D-Mem~\cite{3dmem2024} & 62.9 & 44.7 & 69.1 & 48.9 \\
\midrule
\textbf{HGR (Ours)} & \textbf{64.14} & \textbf{50.1} & \textbf{72.41} & \textbf{56.22} \\
\bottomrule
\end{tabular}
\end{table}

\begin{figure}[t]
\centering
  \includegraphics[width=0.9\linewidth]{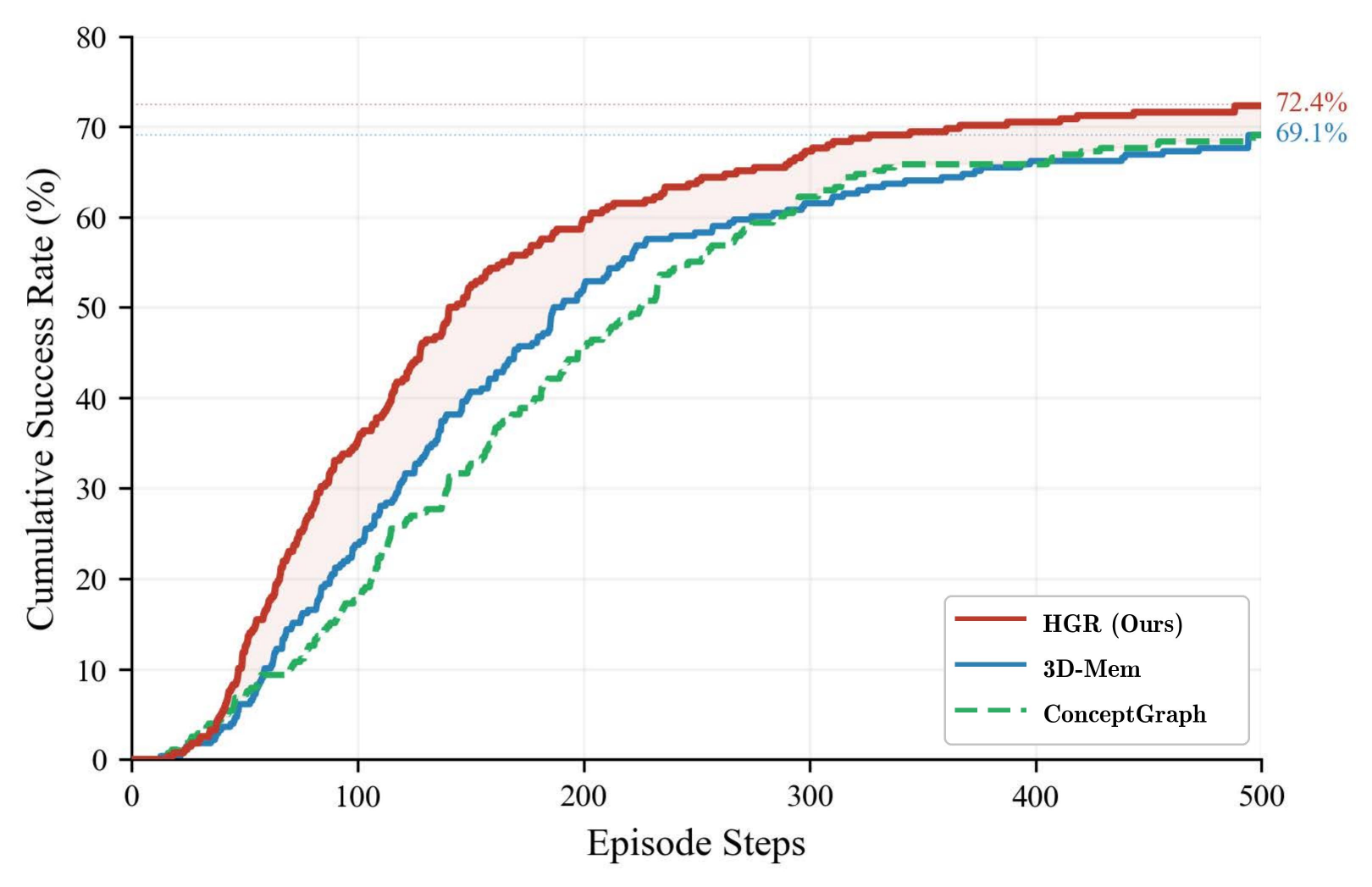}
\caption{\textbf{Cumulative Success Rate vs.\ Episode Steps.} HGR reaches navigation targets earlier than baselines due to hypothesis-driven frontier selection, while 3D-Mem and ConceptGraph require more steps for exhaustive geometric search. The gap widens in later steps as cascade correction prevents error accumulation.}
\label{fig:cumulative_success}
\end{figure}

\paragraph{Where Do the Gains Come From?} As shown in Figure~\ref{fig:cumulative_success}, HGR reaches targets earlier and the advantage grows over longer episodes. Semantic hypothesis module directs exploration toward goal-aligned frontiers, improving SPL, while cascade correction eliminates revisits to erroneously predicted regions.

\paragraph{Modality-Specific Analysis.} Table~\ref{tab:modality} decomposes performance by target modality. HGR shows consistent improvements across all three, with the largest gains on language-specified targets ($+7.9$\% SR over 3D-Mem). Language descriptions benefit most from semantic context propagation, as hypothesis nodes encode relational structure that helps disambiguate spatial references.

\begin{table}[t]
\caption{\textbf{Performance by Target Modality on GOAT-Bench Subset.}}
\label{tab:modality}
\centering
\begin{tabular}{lcccccc}
\toprule
& \multicolumn{2}{c}{Category} & \multicolumn{2}{c}{Language} & \multicolumn{2}{c}{Image} \\
\cmidrule(lr){2-3} \cmidrule(lr){4-5} \cmidrule(lr){6-7}
Method & SR & SPL & SR & SPL & SR & SPL \\
\midrule
ConceptGraph~\cite{conceptgraph2024} & 65.3 & 44.7 & 55.0 & 38.9 & 64.0 & 52.8 \\
3D-Mem~\cite{3dmem2024} & 79.2 & 55.8 & 61.9 & 46.0 & 65.2 & 44.2 \\
\midrule
\textbf{HGR (Ours)} & \textbf{80.9} & \textbf{60.3} & \textbf{69.8} & \textbf{54.5} & \textbf{68.2} & \textbf{61.7} \\
\bottomrule
\end{tabular}
\end{table}

\subsection{Component Ablations}

Table~\ref{tab:ablation} systematically isolates the contributions of each component across all three benchmarks.

\begin{table}[t]
\caption{\textbf{Ablation Study.} Component contributions on the GOAT-Bench subset and question-answering benchmarks. ``Local delete only'' removes the refuted node without cascade propagation along the dependency DAG.}
\label{tab:ablation}
\centering
\small
\begin{tabular}{lccccc}
\toprule
& \multicolumn{2}{c}{GOAT (Subset)} & A-EQA & EM-EQA \\
\cmidrule(lr){2-3} \cmidrule(lr){4-4} \cmidrule(lr){5-5}
Configuration & SR $\uparrow$ & SPL $\uparrow$ & LLM $\uparrow$ & LLM $\uparrow$ \\
\midrule
HGR (full system) & \textbf{72.41} & \textbf{56.22} & \textbf{55.9} & \textbf{58.3} \\
\quad w/o Semantic hypothesis & 67.71 & 50.02 & 48.4 & 50.5 \\
\quad w/o cascade correction & 68.61 & 51.12 & 49.6 & 52.9 \\
\quad Local delete only & 70.85 & 53.67 & 52.3 & 55.9 \\
\quad w/o both (geometry only) & 63.42 & 45.33 & 43.3 & 45.2 \\
\bottomrule
\end{tabular}
\end{table}

\paragraph{Effect of Semantic Hypothesis Module.} Disabling hypothesis projection (reverting to uniform frontier selection) causes consistent degradation: $-4.70$\% SR and $-6.20$\% SPL on GOAT-Bench, $-7.3$ LLM-Match on A-EQA, and $-7.8$ on EM-EQA. This confirms hypothesis H1---semantic guidance substantially reduces exploration redundancy.

\paragraph{Effect of Cascade Correction.} Disabling cascade correction while retaining hypothesis nodes reveals the impact of error accumulation: $-3.80$\% SR and $-5.10$\% SPL on GOAT-Bench as refuted nodes persist with lowered confidence, and $-6.3$ LLM-Match on A-EQA as erroneous scene content influences answers.

\paragraph{Local Delete vs.\ Cascade Correction.} The ``Local delete only'' variant removes the refuted node without propagating along the dependency DAG, achieving 70.85\% SR and 53.67\% SPL---better than no correction but $-$1.56\% SR and $-$2.55\% SPL below the full system. This shows that descendants of false premises degrade performance, and that dependency-aware cascade correction provides gains beyond simple node removal. We also compare against a soft confidence decay baseline with spatial revisitation penalty in Appendix~\ref{app:soft_decay}, confirming that structural deletion outperforms score-based mitigation.

\paragraph{Synergistic Effects.} Disabling both components ($-9.0$\% SR, $-10.9$\% SPL) yields a larger drop than the sum of individual ablations, indicating mutual reinforcement: directed exploration brings the agent to verifiable hypotheses that maximize the correction mechanism's benefit.

\paragraph{Prediction Residual Threshold.} We analyze sensitivity to $\theta_{\text{refute}} \in [0.3, 0.7]$. At $\theta=0.3$ (aggressive), the refutation rate reaches 42\% but incorrectly removes approximately 180 valid nodes, degrading SR by 6.2\%. At $\theta=0.7$ (conservative), only 11\% of hypotheses are refuted, allowing errors to accumulate. The optimal $\theta=0.5$ balances precision (87\%) and recall (74\%).

\paragraph{Hypothesis Prediction Methods.} Table~\ref{tab:prediction_methods} compares VLM-based prediction against heuristic alternatives. An entropy-gated hybrid---VLM for ambiguous frontiers, heuristics otherwise---achieves 95\% of VLM-only SPL at 40\% computational cost. Appendix~\ref{app:open_vlm} further evaluates HGR with open-source VLMs (LLaVA-1.5-13B, InternVL2-8B), confirming consistent gains even with weaker prediction accuracy.

\begin{table}[t]
\caption{\textbf{Hypothesis Prediction Methods.} Trade-offs between prediction accuracy and computational cost on GOAT-Bench subset.}
\label{tab:prediction_methods}
\centering
\small
\begin{tabular}{lccc}
\toprule
Method & Prediction Acc. & Latency (s) & SPL $\uparrow$ \\
\midrule
VLM (GPT-4o) & 0.68 & 0.80 & \textbf{56.22} \\
Heuristic (co-occurrence) & 0.52 & 0.02 & 52.13 \\
Hybrid (entropy-gated) & 0.64 & 0.35 & 55.41 \\
\bottomrule
\end{tabular}
\end{table}

\subsection{Diagnostics: Hypothesis Verification and Correction}

We report statistics on hypothesis lifecycle across the GOAT-Bench full validation set (2,780 subtasks).

\paragraph{Hypothesis Lifecycle.} Over the full evaluation, HGR created 4,712 hypothesis nodes, of which 3,465 (73.5\%) were confirmed upon visitation and transitioned to observed nodes, while 1,247 (26.5\%) were refuted and removed. Of the 1,247 refutations, 342 triggered cascade corrections that removed an average of 2.8 dependent nodes per trigger (maximum cascade depth: 4 hops), totalling approximately 960 descendant removals.

\paragraph{Revisit Reduction.} Cascade correction reduces wasteful revisits to regions based on erroneous predictions. The revisit rate to refuted regions is 4.2\% for HGR, compared to 18.7\% for 3D-Mem. This 4.5$\times$ reduction directly contributes to HGR's SPL improvement.

\paragraph{Error Source Taxonomy.} Among the 1,247 refuted hypotheses: mirror reflections (38\%), glass partitions (29\%), artwork misidentification (18\%), and occlusion artifacts (15\%). Specular and transparent surfaces account for 67\% of all errors, identifying the primary challenge for VLM-based prediction in indoor environments.

\subsection{Secondary Results: A-EQA and EM-EQA}

\paragraph{Active Embodied Question Answering.} Table~\ref{tab:aeqa_results} presents A-EQA results. HGR achieves 55.9 LLM-Match and 45.0 LLM-Match SPL, outperforming all baselines through more efficient exploration toward question-relevant regions and error-free scene memory.

\begin{table}[t]
\caption{\textbf{Performance on A-EQA (184-question subset).} HGR achieves the highest answer quality and exploration efficiency.}
\label{tab:aeqa_results}
\centering
\small
\begin{tabular}{lcc}
\toprule
Method & LLM-Match $\uparrow$ & LLM-Match SPL $\uparrow$ \\
\midrule
\multicolumn{3}{l}{\textit{Blind LLMs (no visual input)}} \\
GPT-4o & 35.9 & --- \\
\midrule
\multicolumn{3}{l}{\textit{Question Agnostic Exploration}} \\
LLaVA-1.5 Frame Captions & 38.1 & 7.0 \\
Multi-Frame & 41.8 & 7.5 \\
\midrule
\multicolumn{3}{l}{\textit{VLM-Guided Exploration}} \\
Explore-EQA & 46.9 & 23.4 \\
ConceptGraph w/ Frontier & 47.2 & 33.3 \\
\textbf{HGR (Ours)} & \textbf{55.9} & \textbf{45.0} \\
\bottomrule
\end{tabular}
\end{table}

\paragraph{Episodic Memory Q\&A.} Table~\ref{tab:emeqa_results} evaluates memory quality in isolation. Given identical pre-explored trajectories, HGR achieves 58.3 LLM-Match with 3.1 average snapshots. Although the exploration path is fixed, HGR still generates hypothesis nodes for visible frontiers as it processes each trajectory step, and verifies them when later steps pass through the predicted regions---so both semantic hypotheses and cascade correction contribute to memory construction quality rather than exploration strategy. The dependency-aware structure enables removal of contradictory information from memory---a capability absent in frame-based representations.

\begin{table}[t]
\caption{\textbf{Performance on EM-EQA.} HGR achieves the highest answer quality with comparable memory footprint.}
\label{tab:emeqa_results}
\centering
\small
\begin{tabular}{lcc}
\toprule
Method & Avg. Frames & LLM-Match $\uparrow$ \\
\midrule
Blind LLM (GPT-4) & 0 & 35.5 \\
ConceptGraph Captions & 0 & 34.4 \\
Frame Captions & 0 & 38.1 \\
\midrule
Multi-Frame & 3.0 & 48.1 \\
\textbf{HGR (Ours)} & \textbf{3.1} & \textbf{58.3} \\
\midrule
Human (Full trajectory) & Full & 86.8 \\
\bottomrule
\end{tabular}
\end{table}

\subsection{Qualitative Analysis}

\begin{figure}[t]
\centering
  \includegraphics[width=\linewidth]{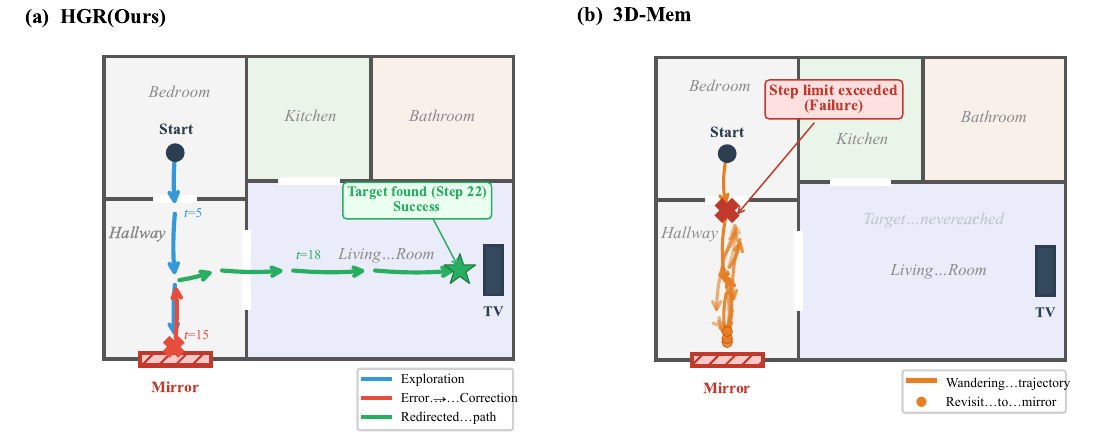}
\caption{\textbf{Qualitative Comparison: Mirror-Induced Prediction Error.} Left: HGR detects the prediction violation and removes the erroneous subgraph via cascade correction. Right: 3D-Mem retains erroneous nodes with lowered confidence, leading to repeated misnavigation.}
\label{fig:qualitative}
\end{figure}

Figure~\ref{fig:qualitative} presents a representative episode searching for a ``television in the living room.'' At step 12, the VLM misidentifies a large mirror as an entryway to an adjacent living room, generating hypothesis nodes for a couch and TV. Upon reaching the mirror (step 15, $\Delta_{\text{sem}} = 0.72 > 0.5$), cascade correction removes the false living room subgraph, redirecting exploration to the actual living room (steps 16--22). In contrast, 3D-Mem retains the erroneous nodes with reduced confidence, causing repeated revisits and step-limit failure.

\section{Conclusion}

We present Hypothesis Graph Refinement (HGR), which addresses a core challenge in embodied navigation: enabling semantic prediction for efficient exploration while preventing cumulative error accumulation. On GOAT-Bench, HGR achieves 72.41\% SR and 56.22\% SPL, outperforming 3D-Mem by +3.31\% and +7.32\% respectively, with consistent improvements on A-EQA and EM-EQA. Ablation studies confirm both mechanisms contribute meaningfully. Diagnostic analysis reveals that 67\% of prediction errors originate from specular and transparent surfaces, identifying a concrete direction for improving VLM robustness in indoor environments. Runtime and memory analysis are provided in Appendix~\ref{app:runtime}.

The main limitation is that cascade correction inherits the accuracy of the prediction residual test: false negatives allow erroneous nodes to persist, while false positives may prune valid hypotheses. More broadly, the prediction quality of HGR is bounded by the underlying VLM, which struggles with visually complex issues such as mirrors, close objects, and spatial relationships between multiple objects—the dominant error source identified in our diagnostics (Appendix~\ref{sec:appendix_fail_cases} provides detailed failure case analysis). Future work includes improving residual calibration to reduce both error modes and extending the dependency structure to dynamic environments where scene content changes over time.


\bibliographystyle{splncs04}
\bibliography{main}

\newpage
\appendix

\section{Detailed Comparison with Commonsense-Guided Exploration}
\label{app:related_comparison}

Table~\ref{tab:related_comparison} compares HGR with recent commonsense-guided exploration methods along five key dimensions. While ESC, SCOPE, and ReVoLT all leverage semantic priors to bias frontier selection, none explicitly models prediction dependencies or provides a mechanism for structural error correction. Tiny Moves proposes game-theoretic hypothesis refinement but does not maintain a persistent dependency graph.

\begin{table}[ht]
\caption{\textbf{Comparison with commonsense-guided exploration methods.}}
\label{tab:related_comparison}
\centering
\small
\begin{tabular}{lccccc}
\toprule
Property & ESC & SCOPE & ReVoLT & Tiny Moves & \textbf{HGR} \\
\midrule
Semantic frontier scoring & \checkmark & \checkmark & \checkmark & \checkmark & \checkmark \\
Persistent graph memory & -- & -- & \checkmark & -- & \checkmark \\
Dependency tracking (DAG) & -- & -- & -- & -- & \checkmark \\
Cascade error correction & -- & -- & -- & -- & \checkmark \\
Non-monotonic graph update & -- & -- & -- & -- & \checkmark \\
\bottomrule
\end{tabular}
\end{table}

ESC applies soft commonsense constraints to weight frontiers during zero-shot object navigation; when a constraint leads to an incorrect direction, the agent must re-explore from scratch without dependency-based propagation.
And SCOPE optimizes semantic scoring of candidate viewpoints for scalable open-vocabulary navigation but treats each prediction independently, so errors in one room's context do not affect subsequent predictions.
ReVoLT combines relational reasoning with Voronoi-based local planning and maintains a local graph, but does not track which predictions depend on which, preventing structural error tracing.
Tiny Moves frames hypothesis refinement as a game-theoretic process with iterative perturbations, complementary to HGR's approach but without a persistent dependency graph across exploration steps.

HGR's key distinction is the dependency DAG, which records derivation relationships among hypotheses. This enables cascade correction---when a parent hypothesis is refuted, all dependent children are automatically identified and removed, preventing downstream error propagation.

\section{Prediction Residual Test Details}
\label{app:residual_details}

\subsection{Verbal Re-statement of Residual Terms}

The prediction residual $\Delta_{\text{sem}}$ (Eq.~3 in the main text) aggregates three complementary signals:

\begin{itemize}
\item \textbf{Category mismatch} $\Delta_c$ (Eq.~4): Measures whether the predicted semantic category matches the actual observation. If the VLM's top-predicted category for a frontier matches the ground-truth category observed upon arrival, $\Delta_c$ is low; otherwise it approaches 1.

\item \textbf{Visual feature divergence} $\Delta_f$ (Eq.~5): Computes $1 - \text{sim}_{\text{CLIP}}$ between the frontier-facing image (captured before visiting) and the actual observation image. This captures visual discrepancies that category labels may miss (e.g., a predicted ``kitchen'' that is actually a laundry room shares similar objects but looks visually different).

\item \textbf{Object-level Jaccard dissimilarity} $\Delta_o$ (Eq.~6): Compares the set of predicted objects $\mathcal{O}_j^{\text{pred}}$ against detected objects $\mathcal{O}_j^{\text{actual}}$ using $1 - \text{IoU}$. This detects cases where the room type is correct but the expected objects are absent.
\end{itemize}

\subsection{Worked Numerical Example}

Consider a frontier where the VLM predicts ``kitchen'' with 0.75 confidence and expects objects \{stove, refrigerator, sink\}. Upon arrival, the agent observes a ``bathroom'' containing \{sink, toilet, bathtub\}.

\begin{itemize}
\item $\Delta_c = 1 - 0.75 \cdot \mathbb{1}[\text{kitchen} = \text{bathroom}] = 1 - 0 = 1.0$
\item $\Delta_f = 1 - \text{sim}_{\text{CLIP}}(I^{\text{frontier}}, I^{\text{actual}}) = 1 - 0.42 = 0.58$
\item $\Delta_o = 1 - \frac{|\{\text{sink}\}|}{|\{\text{stove, refrigerator, sink, toilet, bathtub}\}|} = 1 - 0.2 = 0.80$
\end{itemize}

$\Delta_{\text{sem}} = 0.4 \times 1.0 + 0.3 \times 0.58 + 0.3 \times 0.80 = 0.40 + 0.174 + 0.24 = 0.814$

Since $0.814 > \theta_{\text{refute}} = 0.5$, this hypothesis is refuted and cascade correction is triggered.

\subsection{Per-Term Ablation}

Table~\ref{tab:residual_ablation} reports the effect of dropping each residual term, evaluated on the GOAT-Bench validation subset (278 subtasks).

\begin{table}[t]
\caption{\textbf{Per-term ablation of the prediction residual.} Dropping any single term reduces refutation precision or recall, confirming that all three signals provide complementary information.}
\label{tab:residual_ablation}
\centering
\small
\begin{tabular}{lcccc}
\toprule
Residual Configuration & Precision & Recall & SR $\uparrow$ & SPL $\uparrow$ \\
\midrule
Full ($\Delta_c + \Delta_f + \Delta_o$) & 87\% & 74\% & \textbf{72.41} & \textbf{56.22} \\
Drop $\Delta_c$ (no category) & 78\% & 71\% & 70.88 & 54.31 \\
Drop $\Delta_f$ (no CLIP) & 83\% & 68\% & 71.22 & 55.07 \\
Drop $\Delta_o$ (no object IoU) & 85\% & 66\% & 71.04 & 54.85 \\
\bottomrule
\end{tabular}
\end{table}

Dropping $\Delta_c$ causes the largest precision loss (78\%), as category mismatch is the strongest single indicator of prediction failure. Dropping $\Delta_o$ causes the largest recall loss (66\%), since object-level comparison catches subtle errors where the room type is approximately correct but the expected contents are wrong.

\subsection{Threshold Sensitivity}

We evaluate $\theta_{\text{refute}} \in \{0.3, 0.35, 0.4, 0.45, 0.5, 0.55, 0.6, 0.65, 0.7\}$ on a held-out validation split (56 subtasks from GOAT-Bench).

\begin{table}[t]
\caption{\textbf{Threshold sensitivity for $\theta_{\text{refute}}$.}}
\label{tab:threshold_sensitivity}
\centering
\small
\begin{tabular}{lccccc}
\toprule
$\theta_{\text{refute}}$ & Refutation Rate & Precision & Recall & SR $\uparrow$ & SPL $\uparrow$ \\
\midrule
0.30 & 42\% & 68\% & 91\% & 66.2 & 49.8 \\
0.35 & 37\% & 74\% & 88\% & 68.1 & 51.6 \\
0.40 & 33\% & 79\% & 83\% & 70.0 & 53.4 \\
0.45 & 29\% & 84\% & 78\% & 71.5 & 55.1 \\
\textbf{0.50} & \textbf{24\%} & \textbf{87\%} & \textbf{74\%} & \textbf{72.41} & \textbf{56.22} \\
0.55 & 20\% & 90\% & 67\% & 71.8 & 55.5 \\
0.60 & 16\% & 92\% & 58\% & 70.6 & 53.9 \\
0.65 & 13\% & 93\% & 49\% & 69.2 & 52.1 \\
0.70 & 11\% & 94\% & 41\% & 67.8 & 50.7 \\
\bottomrule
\end{tabular}
\end{table}

The optimal threshold $\theta_{\text{refute}} = 0.5$ balances two failure modes: (1)~aggressive thresholds ($< 0.4$) yield high recall but low precision, incorrectly removing valid hypotheses and degrading SR by up to 6.2\%; (2)~conservative thresholds ($> 0.6$) maintain high precision but allow erroneous hypotheses to persist, leading to cumulative error accumulation. Performance is relatively stable in the $[0.45, 0.55]$ range, indicating moderate sensitivity.

\section{Dependency DAG Construction Rules}
\label{app:dag_construction}

\subsection{Complete Specification}

The dependency DAG $\mathcal{D}$ is constructed incrementally according to the following rules:

\textbf{Rule 1: Frontier-to-Room Hypothesis.} When the VLM generates a room-level semantic prediction for frontier $f_j$, a hypothesis node $v_j^{\text{hyp}}$ is created. The parent edge $(v_{\text{obs}}, v_j^{\text{hyp}}, \rho)$ connects the nearest observed node $v_{\text{obs}}$ to the new hypothesis, where $\rho$ is the VLM's reported confidence for the room prediction.

\textbf{Rule 2: Room-to-Object Hypothesis.} When the VLM further predicts specific objects likely to exist within a hypothesized room (e.g., ``kitchen $\rightarrow$ \{stove, refrigerator, countertop\}''), each predicted object node $v_k^{\text{obj}}$ is created as a child of the room hypothesis node: $(v_j^{\text{hyp}}, v_k^{\text{obj}}, \rho_k)$, where $\rho_k$ is the conditional object probability.

\textbf{Rule 3: Single-Parent Constraint.} Each node has at most one parent in $\mathcal{D}$, ensuring the structure is a forest. If a hypothesis could logically depend on multiple parents (e.g., an object visible from two predicted rooms), we assign it to the parent with the highest confidence $\rho$. This simplifies cascade correction to a simple BFS traversal without requiring multi-parent conflict resolution.

\textbf{Rule 4: Observation Promotion.} When a hypothesis node is confirmed (prediction matches observation), it transitions to an observed node. Its children in $\mathcal{D}$ are preserved but re-parented to the now-verified node, inheriting updated confidence from the observation.

\textbf{Rule 5: No Cycles.} Hypothesis nodes are always created at the frontier of exploration and depend on already-existing nodes. Since exploration proceeds outward, $\mathcal{D}$ is guaranteed to be acyclic.

\subsection{Worked Example}

Consider an agent exploring an apartment. Starting from a hallway (observed node $v_1^{\text{obs}}$):

\begin{enumerate}
\item The agent observes a frontier $f_A$ at the end of the hallway. The VLM predicts ``kitchen'' with $\rho = 0.82$, creating hypothesis node $v_A^{\text{hyp}}$ with edge $(v_1^{\text{obs}}, v_A^{\text{hyp}}, 0.82)$.

\item The VLM further predicts objects in the hypothesized kitchen: ``stove'' ($\rho = 0.71$), ``refrigerator'' ($\rho = 0.68$). Object hypothesis nodes $v_{A1}^{\text{obj}}, v_{A2}^{\text{obj}}$ are created with edges $(v_A^{\text{hyp}}, v_{A1}^{\text{obj}}, 0.71)$ and $(v_A^{\text{hyp}}, v_{A2}^{\text{obj}}, 0.68)$.

\item Another frontier $f_B$ is detected. The VLM predicts ``bedroom'' with $\rho = 0.65$, creating $v_B^{\text{hyp}}$ with edge $(v_1^{\text{obs}}, v_B^{\text{hyp}}, 0.65)$ and child object ``bed'' ($\rho = 0.74$).

\item The agent navigates to $f_A$ and observes a laundry room instead. Since $\Delta_{\text{sem}} = 0.72 > 0.5$, node $v_A^{\text{hyp}}$ is refuted. Cascade correction removes $v_A^{\text{hyp}}$, $v_{A1}^{\text{obj}}$ (stove), and $v_{A2}^{\text{obj}}$ (refrigerator)---three nodes total. The bedroom hypothesis branch ($v_B^{\text{hyp}}$ and its children) is unaffected.
\end{enumerate}

\subsection{Cascade Breadth Analysis}

In practice, dependency chains are shallow. Across the GOAT-Bench evaluation (2,780 subtasks), the distribution of cascade depths was: depth~1 (72\%), depth~2 (21\%), depth~3 (6\%), depth~4 (1\%). The average number of nodes removed per cascade event was 2.8 (including the refuted parent). This bounded cascade breadth ensures that correction operations remain computationally negligible ($< 1$ms per cascade on graphs with $< 20$ nodes).

\section{Open-Source VLM Experiments}
\label{app:open_vlm}

To evaluate whether HGR's gains depend critically on GPT-4o, we replace the VLM model with two open-source alternatives (LLaVA-1.5-13B and InternVL2-8B). Table~\ref{tab:open_vlm} reports results on the GOAT-Bench validation subset (278 subtasks).

\begin{table}[t]
\caption{\textbf{HGR with different VLM backbones on GOAT-Bench subset.} The hypothesis-cascade framework provides consistent gains even with weaker open-source VLMs.}
\label{tab:open_vlm}
\centering
\small
\begin{tabular}{llcccc}
\toprule
& & \multicolumn{2}{c}{Geometry-only} & \multicolumn{2}{c}{HGR (full)} \\
\cmidrule(lr){3-4} \cmidrule(lr){5-6}
VLM Backbone & Pred.\ Acc. & SR & SPL & SR $\uparrow$ & SPL $\uparrow$ \\
\midrule
GPT-4o & 0.68 & 63.42 & 45.33 & \textbf{72.41} & \textbf{56.22} \\
InternVL2-8B & 0.57 & 63.42 & 45.33 & 69.15 & 52.08 \\
LLaVA-1.5-13B & 0.51 & 63.42 & 45.33 & 67.82 & 50.41 \\
\bottomrule
\end{tabular}
\end{table}

Key observations:
\begin{itemize}
\item All VLM backbones improve over the geometry-only baseline, confirming that HGR's framework is not merely a wrapper around GPT-4o's capabilities.
\item The improvement correlates with prediction accuracy: GPT-4o (+8.99\% SR) $>$ InternVL2-8B (+5.73\% SR) $>$ LLaVA-1.5-13B (+4.40\% SR), as expected.
\item Critically, cascade correction becomes \textit{more} important with weaker VLMs: the proportion of hypotheses refuted increases from 26.5\% (GPT-4o) to 35.1\% (LLaVA-1.5-13B), and the relative gain from cascade correction (vs.\ no correction) grows from +3.80\% SR to +5.22\% SR.
\end{itemize}

This demonstrates that HGR is a general framework that benefits from stronger VLMs but does not require them. The cascade correction mechanism is especially valuable when prediction accuracy is lower, as it more aggressively prunes incorrect hypotheses.

\section{Baseline Re-implementation Details}
\label{app:baseline_details}

All baselines share a common infrastructure to ensure fair comparison. Table~\ref{tab:shared_components} details the shared and method-specific components.

\begin{table}[t]
\caption{\textbf{Shared vs.\ method-specific components across all evaluated methods.}}
\label{tab:shared_components}
\centering
\small
\begin{tabular}{lcc}
\toprule
Component & Shared & Method-Specific \\
\midrule
VLM backbone (GPT-4o) & \checkmark & \\
Habitat 3.0 simulator & \checkmark & \\
Low-level controller (DD-PPO) & \checkmark & \\
Occupancy map construction & \checkmark & \\
Geodesic path planner & \checkmark & \\
Step budget (500 steps) & \checkmark & \\
Frontier detection & \checkmark & \\
Graph construction logic & & \checkmark \\
Graph update / correction & & \checkmark \\
Frontier scoring / selection & & \checkmark \\
Memory representation & & \checkmark \\
\bottomrule
\end{tabular}
\end{table}

For ConceptGraph, we re-implement the open-vocabulary scene graph construction using CLIP-based object grounding. The original CLIP backbone is retained for object-node feature extraction, while GPT-4o replaces the original LLM for relational reasoning (edge labeling). Frontier selection uses nearest-frontier with geodesic distance. For 3D-Mem, we re-implement the 3D memory snapshot mechanism; GPT-4o replaces the original VLM for snapshot captioning and query answering, with the top-$k$ relevance retrieval mechanism preserved. For Explore-EQA, the confidence-driven exploration strategy is preserved; GPT-4o serves as both the question-answering module and the exploration confidence estimator.

For all baselines, we verified that our re-implementations match or exceed the originally reported numbers on the respective benchmarks (within 1--2\% variance attributable to VLM backbone differences), confirming faithful reproduction of each method's core logic.

\section{Soft Confidence Decay + Revisitation Penalty Baseline}
\label{app:soft_decay}

A natural alternative to cascade deletion is to \textit{decay} the confidence of suspect hypotheses while adding a spatial revisitation penalty to discourage redundant exploration. We implement this baseline as follows:

\begin{itemize}
\item \textbf{Soft decay}: When a hypothesis node's residual exceeds $\theta_{\text{refute}}$, instead of removing it, multiply its confidence by a decay factor $\gamma = 0.3$. Dependent children have their confidence decayed by $\gamma^d$ where $d$ is the dependency depth.
\item \textbf{Revisitation penalty}: Add a penalty $-\lambda_r \cdot n_{\text{visit}}(f_j)$ to the exploration score (Eq.~2), where $n_{\text{visit}}$ counts prior visits to frontier $f_j$'s Voronoi cell and $\lambda_r = 0.15$.
\end{itemize}

\begin{table}[t]
\caption{\textbf{Cascade correction vs.\ soft confidence decay with revisitation penalty.}}
\label{tab:soft_decay}
\centering
\small
\begin{tabular}{lcccc}
\toprule
& \multicolumn{2}{c}{GOAT (Subset)} & A-EQA & EM-EQA \\
\cmidrule(lr){2-3} \cmidrule(lr){4-4} \cmidrule(lr){5-5}
Method & SR $\uparrow$ & SPL $\uparrow$ & LLM $\uparrow$ & LLM $\uparrow$ \\
\midrule
No correction & 68.61 & 51.12 & 47.5 & 51.8 \\
Soft decay only & 69.42 & 52.30 & 48.8 & 53.1 \\
Soft decay + revisit penalty & 70.17 & 53.05 & 49.6 & 53.9 \\
Local delete only & 70.85 & 53.67 & 52.3 & 55.9 \\
\textbf{Cascade correction (HGR)} & \textbf{72.41} & \textbf{56.22} & \textbf{55.9} & \textbf{58.3} \\
\bottomrule
\end{tabular}
\end{table}

Soft decay with revisitation penalty improves over no correction (+1.56\% SR, +1.93\% SPL) but underperforms both local delete and cascade correction. The key limitation is that decayed nodes remain in the graph and continue to influence downstream reasoning: a ``kitchen'' node decayed to 0.3 confidence still suggests nearby ``stove'' and ``refrigerator'' nodes exist with nonzero probability, biasing exploration toward regions that have already been determined to be incorrectly predicted. Cascade correction eliminates this residual influence entirely.

\section{Full Implementation Details}
\label{app:implementation}

\subsection{Hyperparameters}

\begin{table}[t]
\caption{\textbf{Hyperparameters used in all experiments.}}
\label{tab:hyperparams}
\centering
\small
\begin{tabular}{llc}
\toprule
Symbol & Description & Value \\
\midrule
$\lambda_d$ & Travel cost weight (Eq.~2) & 0.15 \\
$\lambda_h$ & Uncertainty bonus weight (Eq.~2) & 0.10 \\
$r_{\text{context}}$ & Spatial context radius for $\mathcal{O}_{\text{local}}$ & 3.0\,m \\
$\omega_c$ & Category mismatch weight & 0.4 \\
$\omega_f$ & Feature divergence weight & 0.3 \\
$\omega_o$ & Object IoU weight & 0.3 \\
$\theta_{\text{refute}}$ & Refutation threshold & 0.5 \\
$K$ & Number of semantic categories & 23 \\
\bottomrule
\end{tabular}
\end{table}

\subsection{Simulator Configuration}

All experiments use Habitat 3.0 with the following settings:
\begin{itemize}
\item Agent step size: 0.25\,m (forward), $10^{\circ}$ (turn)
\item RGB resolution: $640 \times 480$; depth resolution: $640 \times 480$
\item Maximum episode steps: 500 (GOAT-Bench), 200 (A-EQA), N/A (EM-EQA, pre-explored)
\item Collision handling: agent stops in place and receives collision flag
\item Navigation mesh tolerance: 0.1\,m
\end{itemize}

\subsection{Frontier Detection and Discretization}

Frontiers are detected from the 2D occupancy grid (resolution $0.05$\,m/cell) by identifying cells on the boundary between explored free space and unexplored space. Raw frontier cells are clustered via DBSCAN (eps $= 0.5$\,m, min\_samples $= 3$) to produce discrete frontier regions. Each cluster centroid defines a frontier point $f_j$. On average, 4.7 frontier candidates are identified per timestep across GOAT-Bench episodes.

\subsection{VLM Prompt Template}

The VLM (GPT-4o by default) is prompted with the following template for frontier hypothesis generation:

\begin{verbatim}
You are an embodied navigation assistant. Given:
- Current observation: [image]
- Detected objects nearby: {obj_list}
- Frontier direction: {direction}
- Explored room types so far: {room_list}

Predict the most likely room type beyond this frontier
and list 3-5 objects expected in that room.
Output format:
Room: <room_type> (confidence: 0.0-1.0)
Objects: <obj1> (p=...), <obj2> (p=...), ...
\end{verbatim}

\subsection{Evaluation Protocol}

\begin{itemize}
\item \textbf{GOAT-Bench}: Full validation set (2,780 subtasks across 360 scenes) and a balanced evaluation subset (278 subtasks, 10\% stratified sample). Results are averaged over 3 random seeds.
\item \textbf{A-EQA}: 184-question evaluation subset from 63 HM3D scenes, following the official split. Results averaged over 3 seeds.
\item \textbf{EM-EQA}: 1,634 questions from 152 scenes (ScanNet + HM3D), using provided pre-explored trajectories. Single run (no exploration randomness).
\end{itemize}

\section{Runtime and Computational Cost Analysis}
\label{app:runtime}

\begin{table}[t]
\caption{\textbf{Wall-clock time breakdown per GOAT-Bench episode (avg.\ over 278-subtask evaluation subset).}}
\label{tab:runtime}
\centering
\small
\begin{tabular}{lcc}
\toprule
Component & HGR & 3D-Mem \\
\midrule
Frontier detection + occupancy map & 12\,ms & 12\,ms \\
VLM hypothesis generation (per frontier) & 0.80\,s & --- \\
VLM calls per episode (avg.) & 18.3 & 12.1 \\
Total VLM time per episode & 14.6\,s & 9.7\,s \\
Exploration scoring & 2\,ms & 1\,ms \\
Prediction residual test & 45\,ms & --- \\
Cascade correction (per event) & $<$ 1\,ms & --- \\
Total episode time & 78.4\,s & 63.2\,s \\
\bottomrule
\end{tabular}
\end{table}

HGR adds approximately 24\% wall-clock overhead compared to 3D-Mem, primarily from additional VLM calls for frontier hypothesis generation. The cascade correction itself is negligible ($< 1$\,ms per event: BFS on a graph with typically $< 20$ nodes). The hybrid prediction strategy (Sec.~4.3, Table~5) reduces VLM calls by 60\% while preserving 95\% of performance, offering a practical latency--accuracy trade-off.

When multiple frontiers are discovered simultaneously, VLM queries are batched into a single API call with multiple image inputs, reducing per-frontier latency from 0.80\,s to approximately 0.35\,s for batches of 3--5 frontiers.

\section{Graph Size and Memory Footprint}
\label{app:memory}

\begin{table}[t]
\caption{\textbf{Average graph statistics over GOAT-Bench episodes.}}
\label{tab:graph_stats}
\centering
\small
\begin{tabular}{lccc}
\toprule
Metric & HGR & 3D-Mem & ConceptGraph \\
\midrule
Peak nodes per episode & 34.2 & 41.7 & 38.5 \\
Final nodes per episode & 28.6 & 41.7 & 38.5 \\
Peak edges per episode & 52.1 & 58.3 & 71.2 \\
Nodes removed (cascade) & 5.6 & 0 & 0 \\
Memory footprint (MB) & 8.2 & 12.4 & 15.7 \\
\bottomrule
\end{tabular}
\end{table}

HGR's non-monotonic graph evolution is evident: despite creating more hypothesis nodes during exploration, the final graph size is smaller than baselines due to cascade pruning. The ``grow-and-prune'' dynamics keep the graph compact: peak size occurs around 60\% of the episode, after which cascade corrections increasingly prune erroneous branches while new frontier discoveries slow as the environment is increasingly explored.

The memory footprint advantage (8.2\,MB vs.\ 12.4\,MB for 3D-Mem) arises because: (1) pruned nodes and their associated features are freed from memory, and (2) hypothesis nodes store only lightweight semantic distributions rather than full visual features until confirmed.

\section{Failure Cases Analysis}
\label{sec:appendix_fail_cases}

\begin{figure*}[t]
  \centering
  \begin{subfigure}{0.48\linewidth}
    \includegraphics[width=\linewidth]{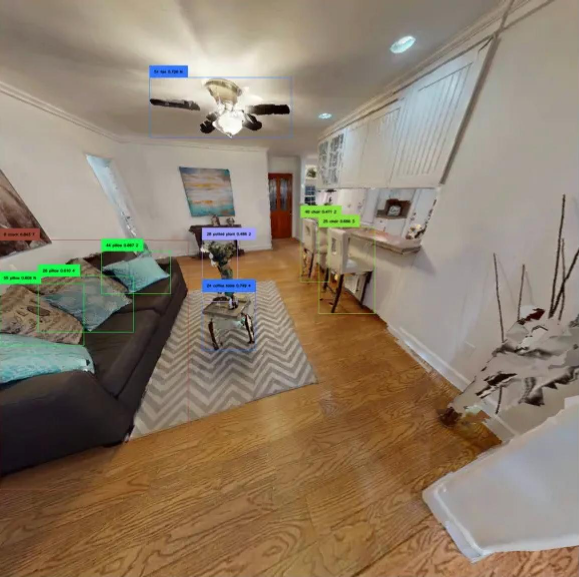}
    \caption{Object Confusion}
    \label{fig:fail_1}
  \end{subfigure}
  \hfill
  \begin{subfigure}{0.48\linewidth}
    \includegraphics[width=\linewidth]{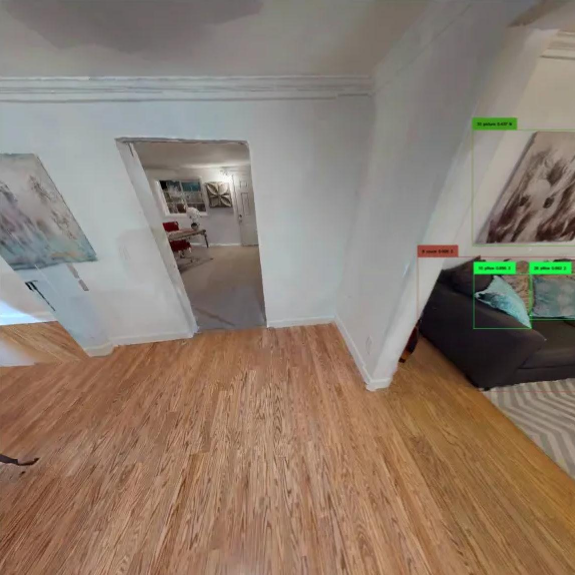}
    \caption{Loss of Details}
    \label{fig:fail_2}
  \end{subfigure}
  
  \vspace{0.4cm}
  
  \begin{subfigure}{0.48\linewidth}
    \includegraphics[width=\linewidth]{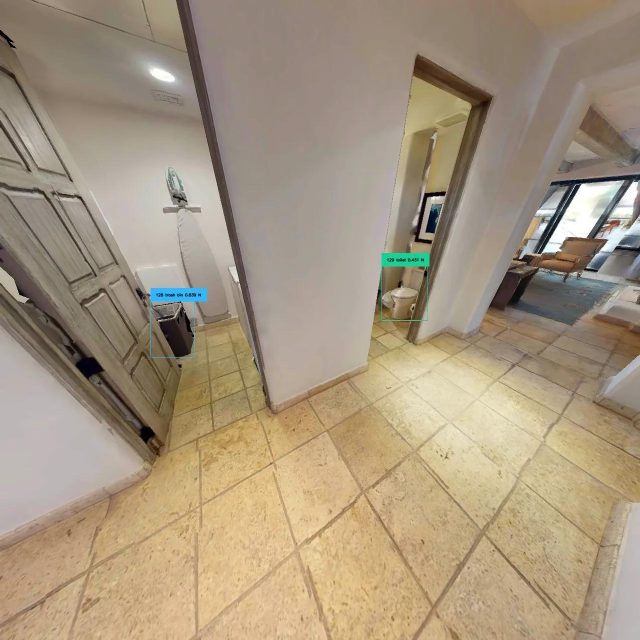}
    \caption{State Misjudgment}
    \label{fig:fail_3}
  \end{subfigure}
  \hfill
  \begin{subfigure}{0.48\linewidth}
    \includegraphics[width=\linewidth]{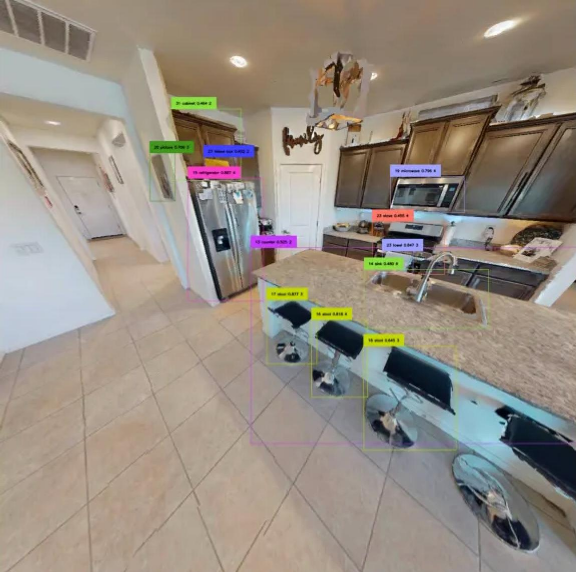}
    \caption{Spatial Misunderstanding}
    \label{fig:fail_4}
  \end{subfigure}
  
  \caption{\textbf{Qualitative Failure Cases of HGR.} We visualize four typical failure patterns: (a) Object confusion caused by detector errors and lack of feature extraction; (b) Loss of visual details due to long observation distances; (c) State misjudgment caused by suboptimal viewpoints or missing subtle visual cues; and (d) Misunderstanding of spatial predicates (e.g., ``between'').}
  \label{fig:fail_cases}
\end{figure*}

While our proposed HGR demonstrates strong performance in 3D scene understanding and reasoning, it still exhibits limitations in certain challenging scenarios. Following the rigorous analysis paradigm in recent 3D embodied models, we thoroughly analyze the typical failure cases and categorize them into four main visual failure modes (as illustrated in Fig.~\ref{fig:fail_cases}), along with an analysis of exploration inefficiency.

\paragraph{Visual Perception and Reasoning Failures.} We identify four primary types of errors during visual grounding and reasoning:
\begin{itemize}
    \item \textbf{Object Confusion (Fig.~\ref{fig:fail_cases}a):} In some cases, the underlying object detector assigns incorrect labels, misleading downstream reasoning. For instance, a wall painting might be erroneously detected as a ceiling fan. The Vision-Language Model (VLM) occasionally fails to extract correct instance-level visual features (e.g., color and shape) to correct this prior error, a problem compounded by missing or inaccurate semantic annotations in the generated scene graph.
    
    \item \textbf{Loss of Visual Details (Fig.~\ref{fig:fail_cases}b):} When the agent observes objects from a considerable distance, fine-grained visual details are easily lost. For example, when queried about the content of a painting above a couch, the VLM might capture low-level textures (e.g., animal fur) but fail to semantically identify the exact subject (e.g., horses) due to insufficient resolution or excessive observation distance.
    
    \item \textbf{State Misjudgment (Fig.~\ref{fig:fail_cases}c):} HGR sometimes struggles with identifying the correct operational states of objects (e.g., whether a toilet seat is open or closed). This typically stems from sub-optimal observation angles, motion blur, or the lack of explicit, state-aware instructions in the prompt (such as ``carefully check the state'') to guide the VLM's attention to subtle visual cues.
    
    \item \textbf{Spatial Relationship Misunderstanding (Fig.~\ref{fig:fail_cases}d):} The model occasionally misinterprets complex spatial predicates (e.g., ``between'', ``left''). For instance, it might incorrectly judge the relative position between a fruit bowl, a knife set, and a toaster. This limitation is primarily attributed to imprecise spatial coordinate mapping within the scene graph and the inherent difficulty VLMs face when grounding 3D spatial semantics from 2D projections.
\end{itemize}

\paragraph{Exploration Inefficiency (SPL).} Beyond visual failures, our quantitative analysis reveals specific cases of path inefficiency where the Success weighted by Path Length (SPL) drops significantly. These inefficiencies typically arise from inaccurate hypothesis node predictions that mislead the agent into exploring erroneous frontiers. Additionally, sub-optimal trade-offs between semantic potential and navigation distance can result in the agent prioritizing distant, low-value frontiers, leading to redundant backtracking. Addressing these memory and navigation trade-offs remains a key direction for future work.

\end{document}